\PassOptionsToPackage{table}{xcolor}
\documentclass{llncs}
\usepackage{algorithm}
\usepackage{algpseudocode}
\usepackage[utf8]{inputenc}
\usepackage{amssymb,amsmath}
\usepackage{scalerel}
\usepackage{graphicx}
\usepackage{tikz}
\usepackage{booktabs}
\usetikzlibrary{shapes.geometric, arrows}
\usepackage{struktex}
\usepackage{gnuplottex}     
\usepackage{etoolbox}
\usepackage{subcaption}
\usepackage{float}
\usepackage{multirow}
\usepackage{cleveref}
\usepackage{tabularx}
\usepackage{multirow}
\usepackage{threeparttable}
\usepackage{lipsum}
\usepackage{indentfirst}
\usepackage{graphicx}
\usepackage[table]{xcolor}
\usepackage{needspace}
\usepackage{multirow}
\AtBeginEnvironment{algorithmic}{\scriptsize}

\crefname{equation}{equation}{equations}
\Crefname{equation}{Equation}{Equations}
\crefrangelabelformat{equation}{(#3#1#4--#5#2#6)}

\crefmultiformat{equation}{(#2#1#3}{, #2#1#3)}{#2#1#3}{#2#1#3}
\Crefmultiformat{equation}{(#2#1#3}{, #2#1#3)}{#2#1#3}{#2#1#3}

\newcommand\Dagger{\scaleobj{1.5}{\dagger}}
\newcommand\DDagger{\scaleobj{1.5}{\ddagger}}
\newcommand\Star{\scaleobj{1.5}{\star}}

\title{Towards Advanced Phenotypic Mutations in Cartesian Genetic Programming}

\author{Roman Kalkreuth} 

\institute{TU Dortmund University, Department of Computer Science, Dortmund, Germany \email{	Roman.Kalkreuth@tu-dortmund.de}}

\begin{document}

{\let\newpage\relax\maketitle}
\setlength{\textfloatsep}{5pt}

\begin{abstract}

Cartesian Genetic Programming is often used with a point mutation as the sole genetic operator. In this paper, we propose two phenotypic mutation techniques and take a step towards advanced phenotypic mutations in Cartesian Genetic Programming. The functionality of the proposed mutations is inspired by biological evolution which mutates DNA sequences by inserting and deleting nucleotides. Experiments with symbolic regression and boolean functions problems show a better search performance when the proposed mutations are in use. The results of our experiments indicate that the use of phenotypic mutations could be beneficial for the use of Cartesian Genetic Programming. 


\end{abstract}

\section{Introduction}

Genetic Programming (GP), as popularized by Koza~\cite{koza1990genetic,koza1992,koza1994genetic}, uses syntax trees as program representation. Cartesian Genetic Programming (CGP) as introduced by Miller et al.~\cite{miller2000cartesian} offers a novel graph-based representation which in addition to standard GP problem domains, makes it easy to be applied to many graph-based applications such as electronic circuits, image processing, and neural networks. CGP is often used only with a point mutation as genetic operator.  
\indent In contrast to tree-based GP for which advanced methods of crossover and mutation have been introduced and investigated, the use of advanced mutation techniques in CGP appears to be mostly unexplored. This significant lack of knowledge in CGP has been the major motivation for our work. Another motivation for our work has been the introduction of a phenotypic subgraph crossover technique for CGP by Kalkreuth et al.~\cite{Kalkreuth2017}. The experiments of Kalkreuth et al. show that the use of the subgraph crossover technique can be beneficial for the search performance of CGP. \\
In standard tree-based GP, the simultaneous use of multiple types of mutation has been found beneficial by Kraft et al.~\cite{Kraft:1994:GPqir} and Angeline et al.~\cite{angeline:1996:leaf}. To our best knowledge, this principle has not been investigated in CGP yet. \par
In this paper, we propose two phenotypic mutations for CGP and take a step towards advanced phenotypic mutations in CGP. Furthermore, we present a first empirical initial testing of both techniques. Section 2 of this paper describes CGP briefly and surveys previous work on phenotypic mutations in CGP. In section 3 we propose our new mutation techniques. Section 4 is devoted to the experimental results and the description of our experiments. In Section 5 we discuss the results of our experiments. Finally, section 6 gives a conclusion and outlines future work.

\section{Related Work}

\subsection{Cartesian Genetic Programming}

Cartesian Genetic Programming is a form of Genetic Programming which offers a novel graph-based representation. In contrast to tree-based GP, CGP represents a genetic program via genotype-phenotype mapping as an indexed, acyclic and directed graph. Originally the structure of the graphs was a rectangular grid of  $N_\textnormal{r}$ rows and $N_\textnormal{c}$ columns, but later work also focused on a representation with at least one row. The genes in the genotype are grouped, and each group refers to a node of the graph, except the last one which represents the outputs of the phenotype. Each node is represented by two types of genes which index the function number in the GP function set and the node inputs. These nodes are called \textit{function nodes} and execute functions on the input values. The number of input genes depends on the maximum arity $N_\textnormal{a}$ of the function set. The last group in the genotype represents the indexes of the nodes which lead to the outputs. A backward search is used to decode the corresponding phenotype. The backward search starts from the outputs and processes the linked nodes in the genotype. In this way, only active nodes are processed during the evaluation.  The number of inputs $N_\textnormal{i}$, outputs $N_\textnormal{o}$ and the length of the genotype is fixed. Every candidate program is represented with $N_\textnormal{r} * N_\textnormal{c} * (N_\textnormal{a} +1) + N_\textnormal{o}$ integers. Even when the length of the genotype is fixed for every candidate program, the length of the corresponding phenotype in CGP is variable which can be considered as a significant advantage of the CGP representation. Figure~\ref{insertion} exemplifies the CGP decoding procedure. \\
\indent CGP traditionally operates within a (1+$\lambda$) evolutionary algorithm (EA) in which $\lambda$ is often chosen with a size of four. The new population in each generation consists of the best individual of the previous population and the $\lambda$ created offspring. The breeding procedure is mostly done by a point mutation which swaps genes in the genotype of an individual in their permissible range by chance. Algorithm~\ref{1+lambda} exemplifies the functioning of the standard (1+$\lambda$)-EA.\\
\indent One of the most important technique is a special rule for the selection of the new parent. In the case when two or more individuals can serve as the parent, an individual which has not served as the parent in the previous generation will be selected as a new parent. This strategy is important because it ensures the diversity of the population and has been found highly beneficial for the search performance of CGP. 

\begin{algorithm}
\caption{Standard (1+$\lambda$)-CGP algorithm}
\begin{algorithmic}[1]
\Procedure{(1+$\lambda$)-CGP}{}
   \State Initialize($P$) \Comment{Initialize parent individual} 
   \State Evaluate($P$) \Comment{Evaluate the fitness of the parent individual} 
   \State Check($P$) \Comment{Check if the parent individual meets the target fitness} 
   \While{true}\Comment{Until termination criteria not triggered}
      \State $Q \gets \textnormal{breed(P)}$ \Comment{Breed $\lambda$ offsprings by mutation} 
      \State Evaluate($Q$) \Comment{Evaluate the fitness of the offsprings} 
      \If {any individual of $Q$ \textit{meets the target fitness}}
      	\State \textbf{return} best individual of $Q$
      \EndIf
      \If {any individual of $Q$ \textit{has better fitness then} $P$}
       	\State Replace $P$ by the offspring with the \textit{best fitness}
      \EndIf
   \EndWhile
\EndProcedure
\end{algorithmic}
\label{1+lambda}
\end{algorithm}

\begin{figure}
\centering
\includegraphics[scale=0.6]{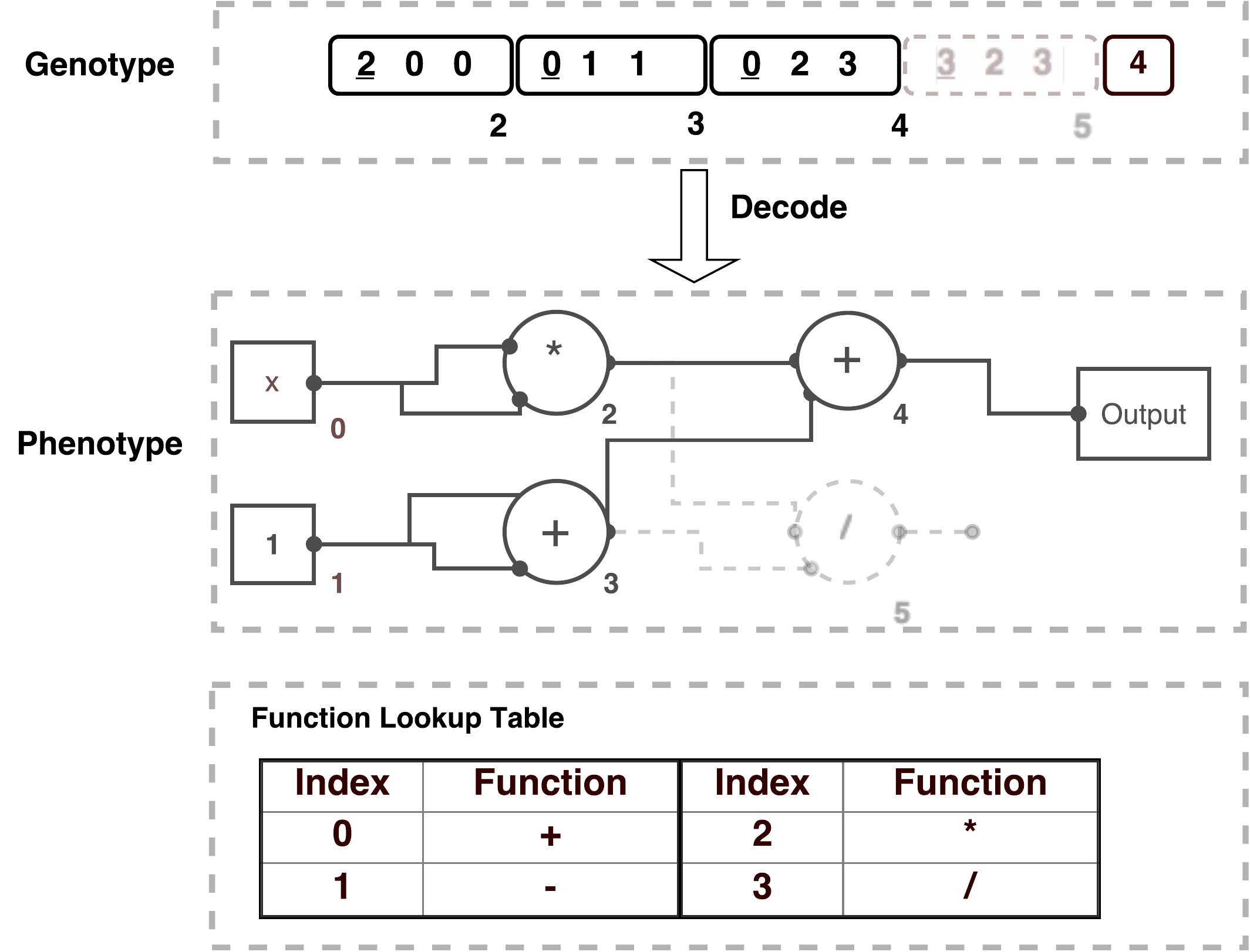}
\caption{Decoding procedure in Cartesian Genetic Programming}
\label{decode_cgp}
\end{figure}   

\newpage 
 
\subsection{Advanced Mutation Techniques in CGP}

For an investigation of the length bias and the search limitation of CGP, a variation of the standard point mutation technique has been introduced by Goldman et al. ~\cite{Goldman:2013:LBS:2463372.2463482}. The modified point mutation mutates exactly one active gene for all offspring. This called \textit{single active-gene mutation strategy} (SAGMS) has been found beneficial for the search performance of CGP. The SAGMS can be seen as a form of phenotypic genetic operator since it respects only active function genes in the genotype which are part of the corresponding phenotype.\\
Later work by Pedroni~\cite{Pedroni} utilized SAGMS to propose an explicitly neutral mutation operator which allows the user to directly control the mutation rate of inactive genes. Pedroni found that the proposed mutation is less sensitive to mutation rate and to reach perfect solutions more consistently than the standard mutation operator.

\section{The Proposed Methods}

The proposed phenotypic mutations for CGP are inspired by biological evolution in which extra base pairs are inserted into a new place in the DNA sequence. This type of mutation is called \textit{insertion}. Another mutation called \textit{deletion} removes sections of the DNA sequence. Related to CGP, we adopt this principle by activating and deactivating randomly chosen function nodes. If a genome is selected for the insertion mutation, one inactive function node becomes active. In the case that all function nodes are active, the individual remains unchanged. Contrary, when deletion is performed, one active node becomes inactive. If the option that outputs can directly connect to the input nodes is disabled, a minimum number of active function nodes has to be defined. The activation and deactivation of the nodes is done by adjusting the connection genes of neighborhood nodes. Both mutation techniques work similarly to the single active-gene mutation strategy. The state of exactly one function node of an individual is changed. Since these forms of mutation can elicit strong changes in the behavior of the individuals, we apply an \textit{insertion rate} and a \textit{deletion rate} for each offspring which is selected for mutation.

\begin{figure}
\centering
\includegraphics[scale=0.8]{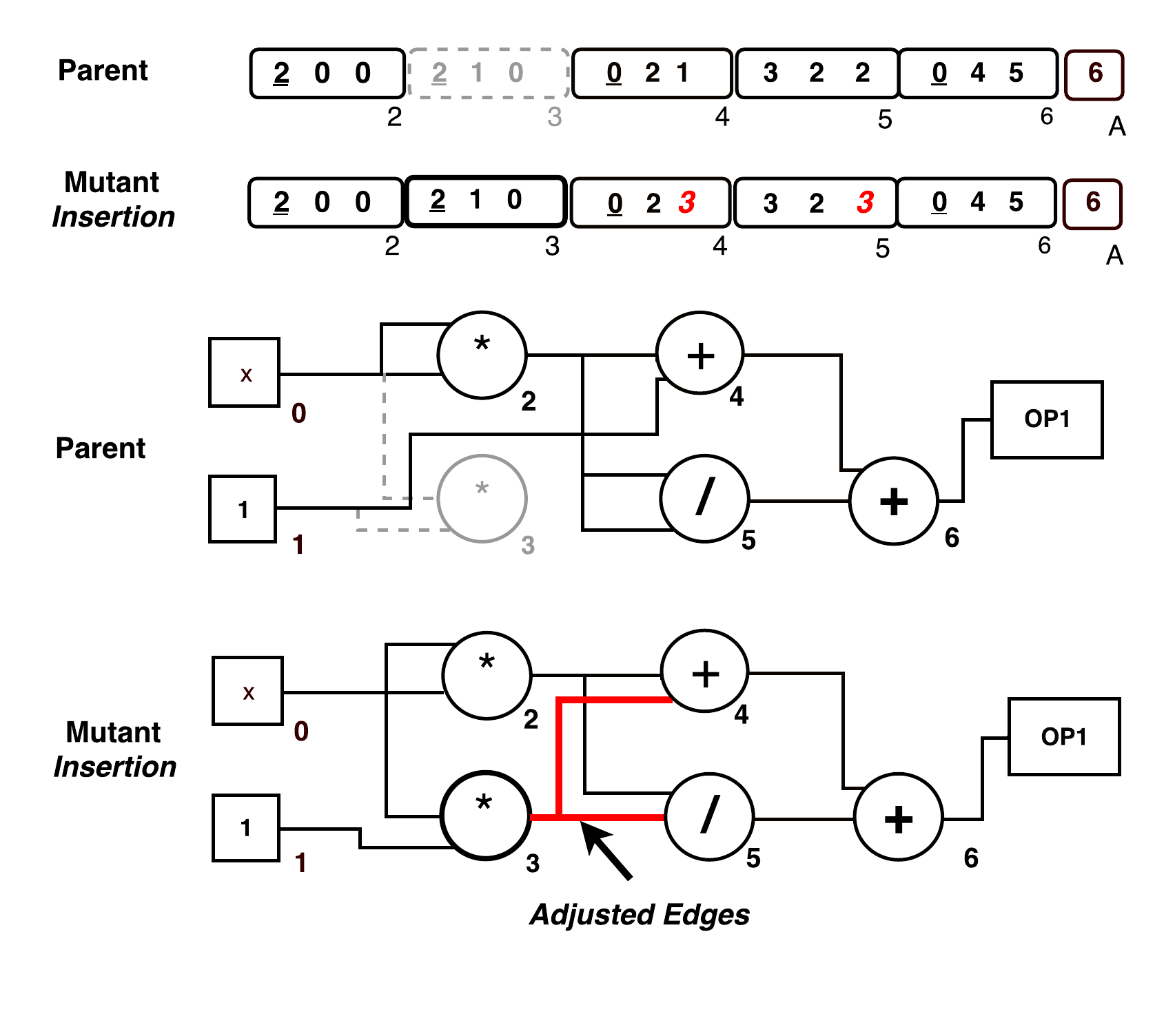}
\caption{The proposed insertion mutation technique}
\label{insertion}
\end{figure}

\begin{figure}
\centering
\includegraphics[scale=0.8]{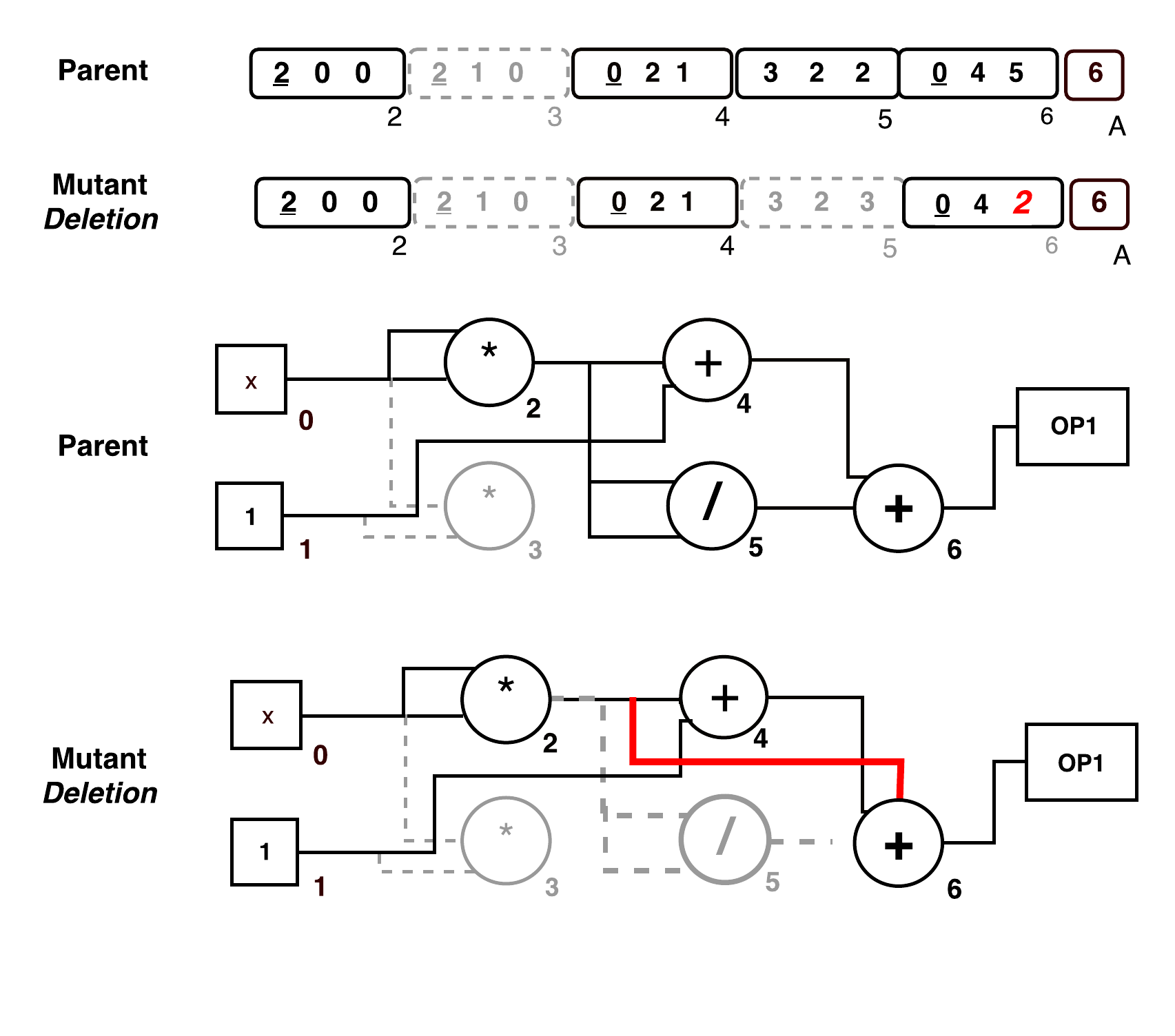}
\caption{The proposed deletion mutation technique}
\label{deletion}
\end{figure}

\indent Fig.~\ref{insertion} and Fig.~\ref{deletion} exemplify the insertion and the deletion mutation techniques. As visible, the connection genes in the genotype are adjusted to deactivate or activate a particular function node in the phenotype.

\newpage

\section{Experiments}

\subsection{Experimental Setup}

We performed experiments on symbolic regression and boolean functions. To evaluate the search performance of the mutations, we measured the number of generations until the CGP algorithm terminated successfully  (\textit{generations-to-success}) and the best fitness value which was found after a predefined number of generations (\textit{best-fitness-of-run}). We used the standard (1+$\lambda$)-CGP algorithm with a population size of 5 individuals ( ($1+4$)-CGP) for our experiments. To classify the significance of our results, we used the Mann-Whitney-U-Test. The mean values are denoted \(a^{\dagger}\) if the $p$-value is less than the significance level $0.05$ and \(a^{\ddagger}\) if the $p$-value is less than the significance level $0.01$ compared to the use of mutation as the sole genetic operator. 
We performed 100 independent runs with different random seeds. \\
\indent For an empirical initial testing of the proposed mutation techniques we defined a range of the mutation ranges from $0 \%$ to $30\%$ inclusive. The setting of important CGP parameters such as point mutation rate has been empirically determined and is oriented with former work. We used the insertion and deletion mutation technique in combination with the standard point mutation technique of CGP. The termination criteria are explained in the particular experiment. For our experiments, we performed a grid analysis of a set of popular benchmark problems which have been used in former GP and CGP experiments. We chose three well known boolean and symbolic regression problems. We analyzed a $4x4$ grid for each problem. The mutations rates for the insertion and deletion mutation techniques are denoted in $10\%$ steps. For instance, a mutation rate of $10 \%$ is denoted as $0.1$. We used a reducing fitness for our experiments. The minimum number of function nodes in the phenotype was set to a number of 4. 

\subsection{Boolean Functions}

For our first experiment in the boolean domain, we chose multiple boolean output problems as the digital adder, subtractor, and multiplier.Former work by White et al.~\cite{gp-benchmarks-2013} proposed these sort of problems as suitable alternatives to the overused parity problems. To evaluate the fitness of the individuals on the multiple output problems, we used compressed truth tables. We defined the fitness value of an individual as the number of different bits. When this number became zero, the algorithm terminated successfully. The configurations for the two-bit boolean functions are shown in Table~\ref{algo_config_mutation_boolean}. In this problem domain, we evaluated the problems with (\textit{generations-to-success}). 

\begin{table}
  \caption{Configuration of the boolean functions problems}
  \setlength{\tabcolsep}{5pt}
  \centering
  \footnotesize
  \scalebox{1.0}{
  \begin{tabular}{ | l | r | r | r |}
    \hline
   \scriptsize \textbf{Property} & \scriptsize \textbf{Add. 2 Bit} & \scriptsize \textbf{Mul. 2 Bit} &\scriptsize  \textbf{Sub. 2 Bit} \\ \hline
     Node count & 30 & 30 & 30 \\ 
     Number of inputs  & 3 & 4 &4 \\	
     Number of outputs & 2 & 4 &3 \\
  	 Function set &  \scriptsize AND, OR & \scriptsize AND, OR & \scriptsize AND, OR, XOR  \\
  	 &  \scriptsize XOR, AND$^{\Star}$  &  \scriptsize  XOR, AND$^{\Star}$ & \scriptsize  NOR, AND$^{\Star}$   \\
 	 Point mutation rate  & 0.05 & 0.05 & 0.05\\
    \hline
    \multicolumn{4}{l}{\textsuperscript{$\Star$}\small{AND with one inverted input}}
  \end{tabular}
  }
  \label{algo_config_mutation_boolean}
\end{table}


\begin{table}
\centering
\begin{tabular}{cc|c|c|c|c|}
\cline{3-6}
& & \multicolumn{4}{ c| }{Insertion Rate} \\ \cline{3-6}
& & \textbf{0.0} & \textbf{0.1} & \textbf{0.2} & \textbf{0.3}\\ \cline{1-6}
\multicolumn{1}{ |c| }{\multirow{4}{*}{Deletion Rate} } &
\multicolumn{1}{ |c| }{\textbf{0.0}} & 86494~~ & $54474^{\DDagger}$ &  $57109^{\DDagger} $ &  $55678^{\DDagger} $   \\ \cline{2-6}
\multicolumn{1}{ |c  }{}   &
\multicolumn{1}{ |c| }{\textbf{0.1}} &  $49710^{\DDagger} $ &  $49894^{\DDagger}$ &  $49767^{\DDagger} $ & $\mathbf{47868^{\DDagger}}$
  \\ \cline{2-6}
\multicolumn{1}{ |c  }{}   &
\multicolumn{1}{ |c| }{\textbf{0.2}} & $\mathbf{36302^{\DDagger}}$ &  $42709^{\DDagger} $ &  $53105^{\DDagger} $ &  $47358^{\DDagger}$ \\ \cline{2-6}
\multicolumn{1}{ |c  }{}   &
\multicolumn{1}{ |c| }{\textbf{0.3}} & $\mathbf{46613^{\DDagger}}$ &  $52043^{\DDagger} $ &  $49382^{\DDagger} $ &  $57263^{\DDagger}$ \\ \cline{1-6}
\cline{3-6}

\end{tabular}
\caption{Mean number of generations for various settings of insertion and deletion mutation for the \textbf{2-Bit digital adder} problem }
\label{adder_2}
\end{table}

\begin{table}
\centering
\begin{tabular}{cc|c|c|c|c|}
\cline{3-6}
& & \multicolumn{4}{ c| }{Insertion Rate} \\ \cline{3-6}
& & \textbf{0.0} & \textbf{0.1} & \textbf{0.2} & \textbf{0.3}\\ \cline{1-6}
\multicolumn{1}{ |c| }{\multirow{4}{*}{Deletion Rate} } &
\multicolumn{1}{ |c| }{\textbf{0.0}} & 17497 & $\mathbf{10995^{\DDagger}}$ & $\mathbf{12290^{\Dagger}} $& $\mathbf{12260^{\Dagger}}$ \\ \cline{2-6}
\multicolumn{1}{ |c  }{}   &
\multicolumn{1}{ |c| }{\textbf{0.1}} & $14682$ & $12873^{\Dagger}$ & $15050~~$ & $14761~~$   \\ \cline{2-6}
\multicolumn{1}{ |c  }{}   &
\multicolumn{1}{ |c| }{\textbf{0.2}} & $18196$ & $14115^{\Dagger}$ & $13440^{\Dagger}$ & $15396~~$  \\ \cline{2-6}
\multicolumn{1}{ |c  }{}   &
\multicolumn{1}{ |c| }{\textbf{0.3}} & $17978$ & $24898~~$ & $15954~~$ & $12414~~$   \\ \cline{1-6}
\cline{3-6}
\end{tabular}
\caption{Mean number of generations for various settings of insertion and deletion mutation for the \textbf{2-Bit multiplier problem} }
\label{mul_2}
\end{table}

\begin{table}
\centering
\begin{tabular}{cc|c|c|c|c|}
\cline{3-6}
& & \multicolumn{4}{ c| }{Insertion Rate} \\ \cline{3-6}
& & \textbf{0.0} & \textbf{0.1} & \textbf{0.2} & \textbf{0.3}\\ \cline{1-6}
\multicolumn{1}{ |c| }{\multirow{4}{*}{Deletion Rate} } &
\multicolumn{1}{ |c| }{\textbf{0.0}} & $14809$ &  $\mathbf{10607^{\DDagger}}$ & $10864^{\DDagger}$ & $11157^{\DDagger}$   \\ \cline{2-6}
\multicolumn{1}{ |c  }{}   &
\multicolumn{1}{ |c| }{\textbf{0.1}} & $14327$ & $11166^{\Dagger}$ & $12079^{\Dagger}$ & $13661~~$   \\ \cline{2-6}
\multicolumn{1}{ |c  }{}   &
\multicolumn{1}{ |c| }{\textbf{0.2}} & $14277$ & $\mathbf{9515^{\DDagger}}$ & $11638^{\Dagger}$  & $12562~~$  \\ \cline{2-6}
\multicolumn{1}{ |c  }{}   &
\multicolumn{1}{ |c| }{\textbf{0.3}} & $15436$ &  $\mathbf{8335^{\DDagger}}$ & $12566^{\Dagger}$ &  $9734^{\DDagger}~$  \\ \cline{1-6}
\cline{3-6}
\end{tabular}
\caption{Mean number of generations for various settings of insertion and deletion mutation for the \textbf{2-Bit subtractor problem} }
\label{sub_2}
\end{table}

Table \ref{adder_2}, \ref{mul_2} and \ref{sub_2} show the results of the grid analysis for the boolean problems. It is clearly visible that the use of the insertion and deletion mutation technique significantly reduces the number of generations until the termination criterion triggered.

\subsection{Symbolic Regression}

For symbolic regression, we have chosen three problems from the work of Clegg et al.~\cite{clegg2007new} and McDermott et al.~\cite{gp-benchmarks-2012} for better GP benchmarks, and the Pagie-1 problem which has been proposed by White et al.~\cite{gp-benchmarks-2013} as an alternative to the heavily overused Koza-1 (``quartic") problem.  The functions of the problems are shown in Table~\ref{problems_regression_mutation}.  A training data set U[$a,b,c$] refers to $c$ uniform random samples drawn from $a$ to $b$ inclusive and E[$a,b,c$] relates to a grid of points evenly spaced with an interval of $c$, from $a$ to $b$ inclusive. The configurations for the problems are shown in Table~\ref{algo_config_mutation_regression}. In this problem domain, we evaluated the problems with the (\textit{best-fitness-of-run}) method. The fitness of the individuals was represented by a cost function value. The cost function was defined by the sum of the absolute difference between the real function values and the values of an evaluated individual. We defined the termination criteria for the experiment with a cost function value less or equal than $0.01$ and a predefined number of generations. We defined a budget of 10000 fitness evaluations for each run. 

\begin{table}
  \caption{Configuration of the symbolic regression problems}
  \setlength{\tabcolsep}{5pt}
  \centering
  \footnotesize
  \scalebox{1.0}{
  \begin{tabular}{ | l | r |}
    \hline
   \scriptsize \textbf{Property} & \scriptsize \textbf{Koza-2,3, Pagie-1} \\ \hline
     Node count & 10  \\ 
     Number of inputs  & 1  \\	
     Number of outputs & 1  \\
  	 Function set &  $+$, $-$, $*$, $/$, $\sin$, $\cos$,  $\ln(|n|)$, $e^n$ \\
 	 Point mutation rate  & 0.2\\
    \hline
  \end{tabular}
  }
  \label{algo_config_mutation_regression}
\end{table}

\begin{table}
  \caption{Symbolic regression problems of the first experiment}
  \setlength{\tabcolsep}{5pt}
  \footnotesize
  \centering
  \scalebox{1.0}{
  \begin{tabular}{ | l | l | r | l |}
    \hline
   \textbf{Problem} & \textbf{Objective Function} & \textbf{Vars} &\textbf{Training Set}\\ \hline
    Koza-2	&  $ x^5 - 2x^3 +x$ & 1 & U[-1,1,20]\\
	Koza-3	&  $ x^6 - 2x^4 +x^2$ & 1 & U[-1,1,20]\\
	Pagie-1 & $1/(1+x^{-4})+1/(1+y^{-4})$ & 2 &  E[-5,5,0.4]\\
     \hline
  \end{tabular}
  }
  \label{problems_regression_mutation}
  \end{table}

  
\begin{table}
\centering
\begin{tabular}{cc|c|c|c|c|}
\cline{3-6}
& & \multicolumn{4}{ c| }{Insertion Rate} \\ \cline{3-6}
& & \textbf{0.0} & \textbf{0.1} & \textbf{0.2} & \textbf{0.3}\\ \cline{1-6}
\multicolumn{1}{ |c| }{\multirow{4}{*}{Deletion Rate} } &
\multicolumn{1}{ |c| }{\textbf{0.0}} & $0,33~~$ &  $0,20^{\DDagger}$ & $\mathbf{0,18^{\DDagger}}$ & $\mathbf{0,17^{\DDagger}}$ \\ \cline{2-6}
\multicolumn{1}{ |c  }{}   &
\multicolumn{1}{ |c| }{\textbf{0.1}} & $0,25~~$ & $0,21^{\DDagger}$ &  $\mathbf{0,17^{\DDagger}}$ & $0,20^{\DDagger}$  \\ \cline{2-6}
\multicolumn{1}{ |c  }{}   &
\multicolumn{1}{ |c| }{\textbf{0.2}} & $0,25~~$  & $0,20^{\DDagger}$ & $\mathbf{0,18^{\DDagger}}$ & $0,19^{\DDagger}$  \\ \cline{2-6}
\multicolumn{1}{ |c  }{}   &
\multicolumn{1}{ |c| }{\textbf{0.3}} & $0,23~~$ & $0,21^{\DDagger}$ & $0,21^{\DDagger}$ & $0,21^{\DDagger}$  \\ \cline{1-6}
\cline{3-6}
\end{tabular}
\caption{Mean best fitness of run for various settings of insertion and deletion mutation for the \textbf{Koza-2} problem}
\label{Koza2}
\end{table}


\begin{table}
\centering
\begin{tabular}{cc|c|c|c|c|}
\cline{3-6}
& & \multicolumn{4}{ c| }{Insertion Rate} \\ \cline{3-6}
& & \textbf{0.0} & \textbf{0.1} & \textbf{0.2} & \textbf{0.3}\\ \cline{1-6}
\multicolumn{1}{ |c| }{\multirow{4}{*}{Deletion Rate} } &
\multicolumn{1}{ |c| }{\textbf{0.0}} & $0,34~~$ &  $0,25^{\DDagger}$ & $\mathbf{0,23^{\DDagger}}$ & $\mathbf{0,22^{\DDagger}}$ \\ \cline{2-6}
\multicolumn{1}{ |c  }{}   &
\multicolumn{1}{ |c| }{\textbf{0.1}} & $0,32~~$ & $0,24^{\DDagger}$ &  $\mathbf{0,22^{\DDagger}}$ & $0,23^{\DDagger}$  \\ \cline{2-6}
\multicolumn{1}{ |c  }{}   &
\multicolumn{1}{ |c| }{\textbf{0.2}} & $0,30~~$  & $0,26^{\DDagger}$ & $0,24^{\DDagger}$ & $0,23^{\DDagger}$ \\ \cline{2-6}
\multicolumn{1}{ |c  }{}   &
\multicolumn{1}{ |c| }{\textbf{0.3}} & $0,32~~$ & $0,26~~
$ & $0,24^{\DDagger}$ & $0,22^{\DDagger}$  \\ \cline{1-6}
\cline{3-6}
\end{tabular}
\caption{Mean best fitness of run for various settings of insertion and deletion mutation for the \textbf{Koza-3} problem}
\label{Koza3}
\end{table}

\begin{table}
\centering
\begin{tabular}{cc|c|c|c|c|}
\cline{3-6}
& & \multicolumn{4}{ c| }{Insertion Rate} \\ \cline{3-6}
& & \textbf{0.0} & \textbf{0.1} & \textbf{0.2} & \textbf{0.3}\\ \cline{1-6}
\multicolumn{1}{ |c| }{\multirow{4}{*}{Deletion Rate} } &
\multicolumn{1}{ |c| }{\textbf{0.0}} & $195,90~~
$ &  $186,12~~$ & $\mathbf{176,23^{\DDagger}}$ & $\mathbf{175,45^{\DDagger}}$ \\ \cline{2-6}
\multicolumn{1}{ |c  }{}   &
\multicolumn{1}{ |c| }{\textbf{0.1}} & $194,12~~$ & $185,55^{\Dagger}$ &  $\mathbf{175,03^{\DDagger}}$ & $172,52^{\DDagger}$  \\ \cline{2-6}
\multicolumn{1}{ |c  }{}   &
\multicolumn{1}{ |c| }{\textbf{0.2}} & $192,69~~$  & $183,32^{\Dagger}$ & $187,43~~$ & $179,07^{\DDagger}$ \\ \cline{2-6}
\multicolumn{1}{ |c  }{}   &
\multicolumn{1}{ |c| }{\textbf{0.3}} & $193,93~~$ & $191,85~~
$ & $183,04^{\Dagger}$ & $178,39^{\DDagger}$  \\ \cline{1-6}
\cline{3-6}
\end{tabular}
\caption{Mean best fitness of run for various settings of insertion and deletion mutation for the \textbf{Pagie-1} problem}
\label{Pagie1}
\end{table}

Table \ref{Koza2}, \ref{Koza3} and \ref{Pagie1} show the results of the grid analysis for the symbolic regression problems. It is clearly visible that the use of the insertion and deletion mutation technique significantly improves the fitness after a predefined number of fitness evaluations.

\section{Discussion}

Our experiments indicate that the \textit{insertion} and \textit{deletion} mutation may be beneficial for the search performance of CGP. Furthermore, our experiments showed beneficial effects in two different problem domains. The results of our experiments also show that the proposed phenotypic mutations can significantly contribute to the search performance for different types of fitness. However, in our experiments, we only investigated simple test problems which are well known in the field of GP and CGP. For more significant statements about the potential of the \textit{insertion} and \textit{deletion} mutation techniques in CGP a more detailed and comprehensive study is needed. This study should include a larger set of \textit{state-of-the-art} benchmarks which have been proposed by McDermott et al.~\cite{gp-benchmarks-2012} and White et al.~\cite{gp-benchmarks-2013} Furthermore, this study should also include more problem domains. Beside to the question of contributions to the search performance of CGP, it has to be investigated in which way the proposed mutations contribute to the search performance. A suitable method to investigate the behavior and effect of both mutations would be an analysis of the exploration abilities of the search space. A method which could contribute to achieving more insight into the exploration behavior of both mutations would be an analysis of the fitness landscape. This analysis could also answer the question if our proposed mutations contribute to the overstepping of local optima. \\
\indent Another important point to discuss is the parametrization of the \textit{insertion} and \textit{deletion} mutation technique. On some problems, the simultaneous use of both mutations showed beneficial effects. However, our experiments also show that the sole use of one phenotypic mutation can also beneficial for the search performance. Therefore we think that a detailed investigation of different cases is needed in which the use of both or only one mutation is beneficial.\\
\indent The last point which should be discussed is the complexity of the \textit{insertion} and \textit{deletion} mutation. To activate and deactivate certain function nodes, both mutations require a permanent listing of the active function nodes and the corresponding structure of connections. Therefore we think that the runtime of both mutations should be investigated on a theoretical and practical level. The results of the runtime measurement should also be compared to the runtime of the respective fitness evaluation tasks. This type of comparison could be very helpful to get more clearness if the \textit{insertion} and \textit{deletion} mutation are really beneficial for the search performance of CGP.

\section{Conclusion and Future Work}

Within this paper, we proposed two new phenotypic mutations and took a step towards advanced phenotypic mutations in CGP. A first initial empirical testing indicates that the use of both mutation techniques could be beneficial for the search performance of CGP. Our experiments clearly show that the use of the \textit{insertion} and \textit{deletion} mutation techniques can significantly improve the search performance of CGP for our tested problems. However, for more significant statements about the beneficial effects of the proposed mutations, a rigorous and comprehensive study on a larger set of problems is needed and should include the investigation of different CGP algorithms. Consequently, we will mainly focus on more detailed and comprehensive experiments in the future. These experiments will also include an analysis of the exploration abilities of CGP when the proposed mutations are in use. Another part of our future work is devoted to a detailed investigation of the parametrization of both mutation techniques. This will also include investigations in which way both mutations work together and if there are similar functional behaviors between different problems.

\bibliographystyle{splncs03}
\bibliography{literature}

\end{document}